\documentclass[conference]{IEEEtran}
\IEEEoverridecommandlockouts

\usepackage{amsmath,amsfonts}
\usepackage{algorithm}
\usepackage{algpseudocode}
\usepackage{array}
\usepackage{textcomp}
\usepackage{stfloats}
\usepackage{graphicx}
\usepackage{cite}
\usepackage{multirow}
\usepackage{hyperref} 
\usepackage{float}

\usepackage{wrapfig}
\usepackage{tikz}

\begin{document}

\title{Efficient Client Selection in Federated Learning}
\author{\IEEEauthorblockN{William Marfo, Deepak K. Tosh, Shirley V. Moore} \\
\IEEEauthorblockA{\textit{Department of Computer Science}, \textit{University of Texas at El Paso}, El Paso, USA \\
\ wmarfo@miners.utep.edu, dktosh@utep.edu, svmoore@utep.edu
}}

\maketitle

\begin{abstract}
Federated Learning (FL) enables decentralized machine learning while preserving data privacy. This paper proposes a novel client selection framework that integrates differential privacy and fault tolerance. The adaptive client selection adjusts the number of clients based on performance and system constraints, with noise added to protect privacy. Evaluated on the UNSW-NB15 and ROAD datasets for network anomaly detection, the method improves accuracy by 7\% and reduces training time by 25\% compared to baselines. Fault tolerance enhances robustness with minimal performance trade-offs.
\end{abstract}

\footnote{This material is based upon work supported by the United States Department of Energy’s (DOE) Office of Fossil Energy (FE) Award DE-FE0031744.}


\section{Introduction}
Federated Learning (FL) enables distributed machine learning without centralizing data, addressing critical privacy concerns \cite{marfo2022network}. However, efficient client selection and robust privacy preservation mechanisms remain key challenges in FL \cite{li2024adafl}. Poor client selection can degrade model performance, while insufficient privacy safeguards may expose sensitive information. Moreover, heterogeneity in client data and system resources can lead to bottlenecks, slowing down training. In this paper, we propose an efficient client selection method that integrates differential privacy (DP) and fault tolerance mechanisms to enhance model performance and system resilience. The adaptive selection process adjusts the number of clients dynamically based on performance and system constraints, with added noise ensuring data privacy \cite{ li2024adafl}. To evaluate our framework, we apply it to the UNSW-NB15 and ROAD datasets for network anomaly detection \cite{moustafa2015unsw, verma2024comprehensive}, demonstrating significant improvements in accuracy and training efficiency compared to baselines.

\section{Proposed Client Selection Method}
\label{sec:Proposed}

We propose a client selection method for FL that balances accuracy, privacy, and fault tolerance. The method selects a subset of clients based on their potential contribution to the global model while incorporating differential privacy (DP) and fault tolerance mechanisms. Our approach has two main components: (1) an adaptive client selection algorithm and (2) the integration of DP and checkpointing.

In each round, available clients \( A_t \) are evaluated based on utility scores, which are computed using factors such as data quality and computational capacity. The top \( K \) clients are selected to train local models. Differential privacy is ensured by adding Gaussian noise to the model updates, controlled by the privacy budget \( \epsilon \), to prevent the server from inferring individual client data during aggregation. To enhance fault tolerance, a checkpointing mechanism, with intervals determined by \( t_c^* \), allows clients to save and recover from failures during training. Algorithm~\ref{alg:client_selection} presents the proposed method, outlining the client selection, privacy protection, and fault tolerance procedures.

\begin{algorithm}
\caption{Client selection with differential privacy and fault tolerance}
\label{alg:client_selection}
\begin{algorithmic}[1]
\Require Set of all clients \( C \), number of clients to select \( K \), privacy budget \( \epsilon \), checkpointing interval \( t_c^* \)
\Ensure Selected subset of clients \( S_t \) for each round \( t \)
\State Initialize utility scores for all clients
\For{each round \( t \)}
    \State Get available clients \( A_t \)
    \State Compute utility scores for each client in \( A_t \)
    \State Select top \( K \) clients based on utility scores
    \For{each selected client}
        \State Train local model, apply gradient clipping
        \State Add Gaussian noise to gradients for DP
        \State Send noisy gradients to the server
        \If{checkpoint interval reached} Save checkpoint
        \EndIf
        \If{client failure detected} Recover from checkpoint
        \EndIf
    \EndFor
    \State Aggregate updates and update global model
\EndFor
\end{algorithmic}
\end{algorithm}

\section{Performance Evaluation}
\label{sec:Evaluation}

\subsection{Experimental Setup}
\label{sec:Experimental}

We used two widely recognized network security datasets: the UNSW-NB15 \cite{moustafa2015unsw} and ROAD \cite{verma2024comprehensive}. The experiments were conducted on a system with a 12th Gen Intel Core i9-12900HK, NVIDIA RTX 3080 Ti GPU, and 32GB of RAM, using \texttt{Python 3.8}, \texttt{TensorFlow 2.6.0}, and \texttt{PyTorch 0.5.0}. For baseline comparisons, we used:
- \textbf{ACFL} \cite{yan2023criticalfl}: an active learning-based client selection method.
- \textbf{FedL2P} \cite{lee2024fedl2p}: a meta-learning approach for personalized fine-tuning.

Evaluation metrics include accuracy and AUC-ROC, which are standard for anomaly detection tasks.

\subsection{Results and Analysis}
\label{sec:Results}

\subsubsection{Performance Comparison with Baselines}

\begin{figure}[h]
\centering
\includegraphics[width=0.43\textwidth]{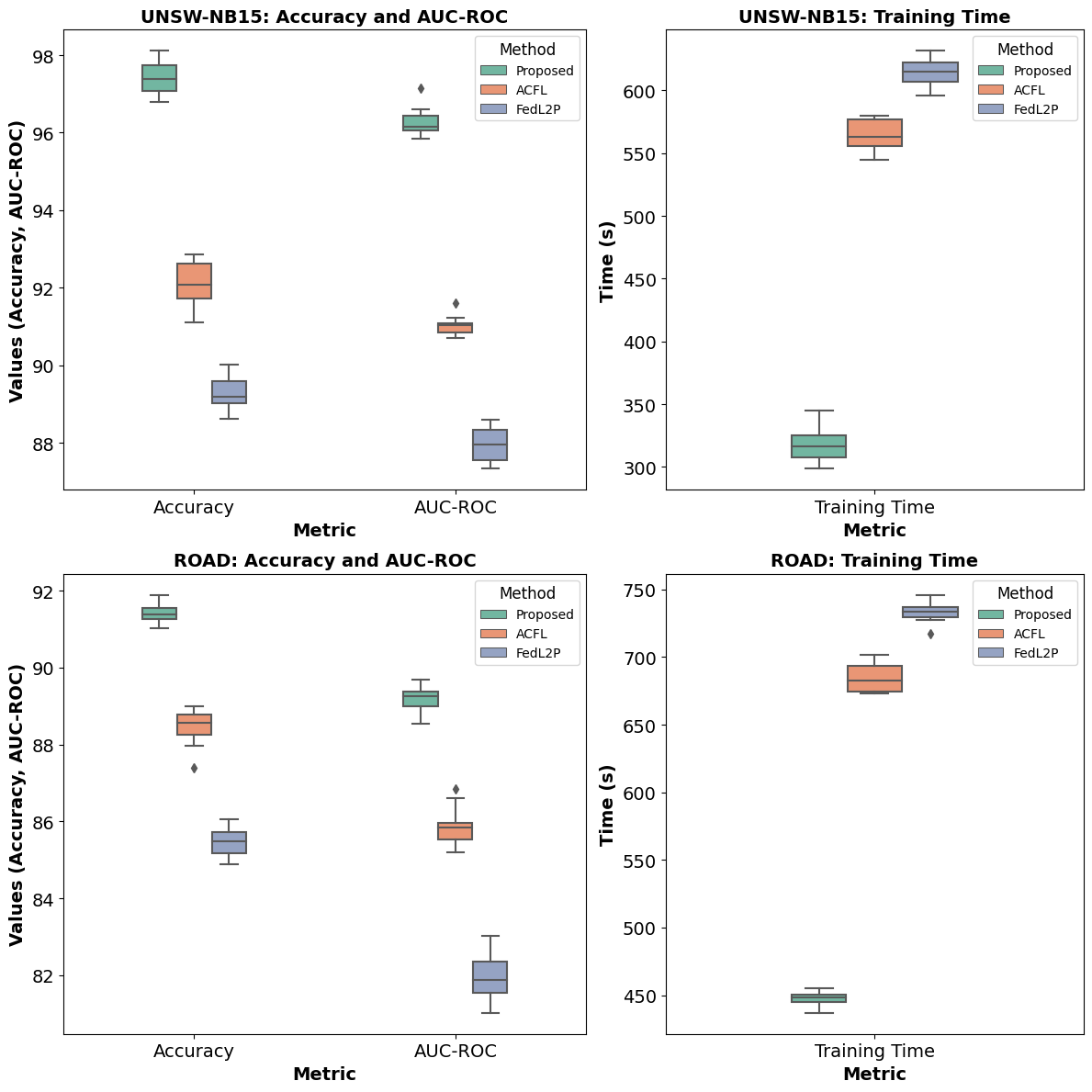}
\caption{Performance comparison of the proposed method, ACFL, and FedL2P in terms of accuracy, AUC-ROC, and training time.}\vspace{-1em}
\label{fig:performance_comparison}
\end{figure}

We evaluated the performance of our method against two baseline approaches, ACFL and FedL2P. As depicted in Fig.~\ref{fig:performance_comparison}, our proposed method demonstrates a 7\% improvement in accuracy and a 25\% reduction in training time compared to the baselines. These gains are particularly pronounced in the ROAD dataset, where our method effectively handles complex anomaly patterns, showcasing its robustness in diverse network conditions.

\subsubsection{Impact of Differential Privacy}

\begin{figure}[t]
\centering
\includegraphics[width=0.48\textwidth]{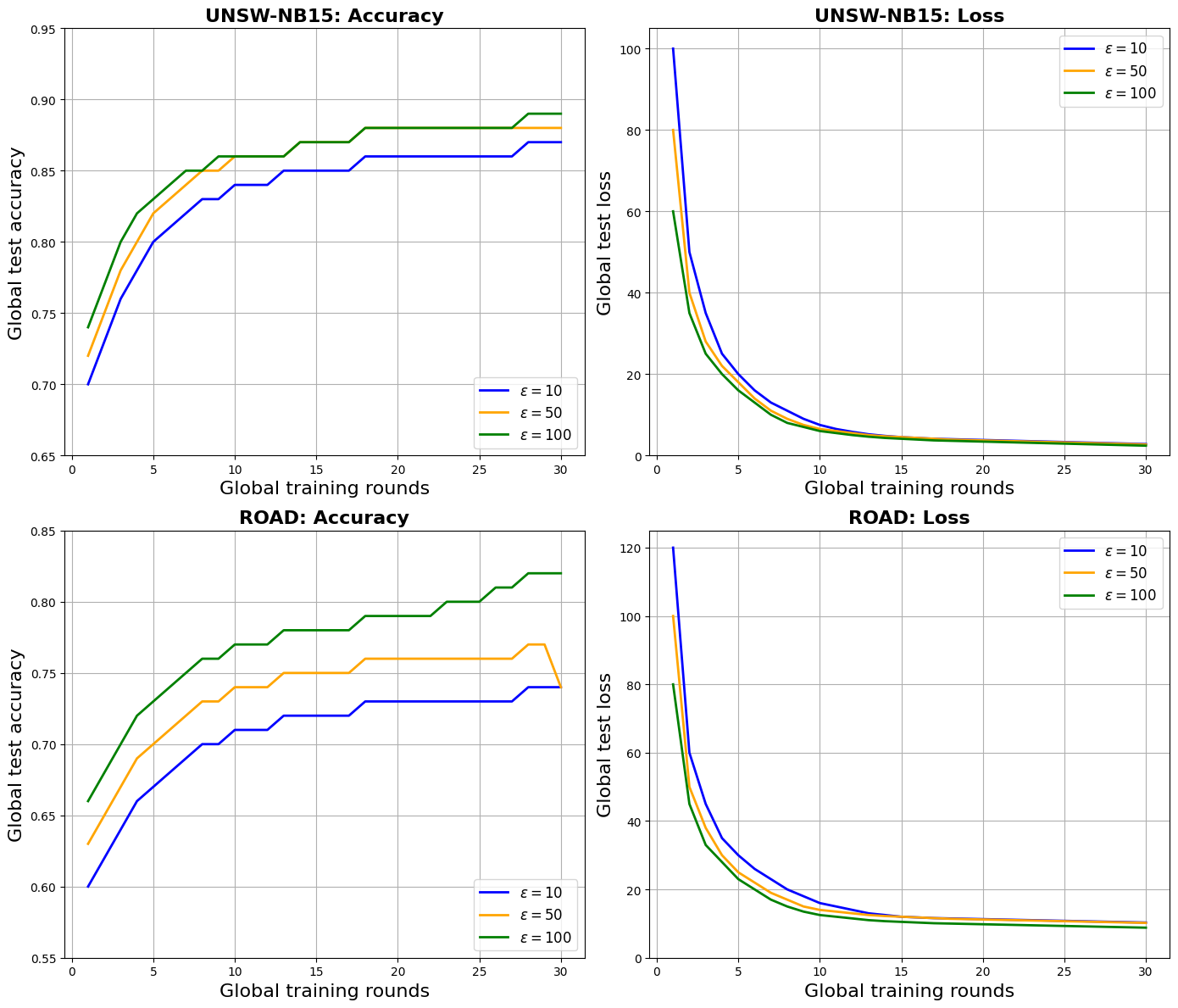}
\caption{Impact of privacy budgets on accuracy and loss for UNSW-NB15 and ROAD datasets.}
\label{fig:dp_impact}
\end{figure}

We analyzed the effect of different privacy budgets (\(\epsilon\)) on model performance. Fig.~\ref{fig:dp_impact} shows that increasing the privacy budget leads to improved accuracy and reduced loss across both datasets. For instance, in the UNSW-NB15 dataset, accuracy increased from 86\% at \(\epsilon = 10\) to 89\% at \(\epsilon = 100\), while similar trends were observed in the ROAD dataset, where accuracy improved from 73\% to 82\%. These results demonstrate the balance between maintaining privacy and ensuring model performance.

\subsubsection{Effect of Fault Tolerance}

We evaluated the impact of introducing fault tolerance mechanisms through checkpointing. Table~\ref{tab:fault_tolerance_impact} shows that while accuracy and AUC-ROC experienced a slight decline (approximately 2-3\%), the overall system robustness improved by effectively handling client dropouts. Training time increased by around 5-10\%, but this is an acceptable trade-off given the enhanced system reliability.

\begin{table}[h]
\centering
\caption{Impact of fault tolerance on model performance (UNSW-NB15 and ROAD).}
\label{tab:fault_tolerance_impact}
\begin{tabular}{l|c|c|c}
\hline
\textbf{Configuration} & \textbf{Accuracy (\%)} & \textbf{AUC-ROC} & \textbf{Training Time (s)} \\ \hline
\multicolumn{4}{c}{\textbf{UNSW-NB15}} \\ \hline
Without Fault Tolerance & 94.8 & 0.93 & 570 \\
With Fault Tolerance    & 92.1 & 0.91 & 600 \\ \hline
\multicolumn{4}{c}{\textbf{ROAD}} \\ \hline
Without Fault Tolerance & 90.3 & 0.88 & 680 \\
With Fault Tolerance    & 88.7 & 0.86 & 710 \\
\hline
\end{tabular}
\end{table}

\section{Conclusion}
\label{sec:conclusion}

We proposed an efficient client selection method for FL that integrates differential privacy and fault tolerance, validated on network anomaly detection tasks. Our method improved accuracy by 7\% and reduced training time by 25\% compared to FedL2P, demonstrating a balance between privacy, performance, and robustness. Future work will explore adaptive hyperparameter tuning and other privacy-preserving techniques.

\bibliographystyle{IEEEtran} 
\bibliography{conference_101719} 

\end{document}